\documentclass{article}

 \PassOptionsToPackage{square,numbers,sort&compress}{natbib}
\usepackage[preprint]{neurips_2019_arxiv}
\usepackage[utf8]{inputenc} % allow utf-8 input
\usepackage[T1]{fontenc}    % use 8-bit T1 fonts
\usepackage{hyperref}       % hyperlinks
\usepackage{url}            % simple URL typesetting
\usepackage{booktabs}       % professional-quality tables
\usepackage{amsfonts}       % blackboard math symbols
\usepackage{nicefrac}       % compact symbols for 1/2, etc.
\usepackage{microtype}      % microtypography
\usepackage{amsmath}  % Custom
\usepackage{enumitem}  % Custom
\usepackage{pdfpages} % Custom - for appendix
\usepackage{booktabs}  %Required
\usepackage{caption}

\DeclareMathOperator*{\argmax}{arg\text{ }max}
\newcommand{\sig}{\textsuperscript{$\dagger$}\hspace{0.0ex}}
\newcommand{\isep}{\mathrel{{.}\,{.}}\nobreak}

\title{Learning Counterfactual Representations for Estimating Individual Dose-Response Curves}

\author{
  Patrick Schwab$^{1}$, Lorenz Linhardt$^{2}$, Stefan Bauer$^{3}$, Joachim M. Buhmann$^{2}$, Walter Karlen$^{1}$ \\
  $^{1}$Institute of Robotics and Intelligent Systems, ETH Zurich, Switzerland \\
  $^2$ Department of Computer Science, ETH Zurich, Switzerland \\
  $^3$ MPI for Intelligent Systems, Tübingen, Germany\\
  \texttt{patrick.schwab@hest.ethz.ch} \\
}

\begin{document}

\maketitle
\vskip -3.5ex

\begin{abstract}
Estimating what would be an individual's potential response to varying levels of exposure to a treatment is of high practical relevance for several important fields, such as healthcare, economics and public policy. However, existing methods for learning to estimate counterfactual outcomes from observational data are either focused on estimating average dose-response curves, or limited to settings with only two treatments that do not have an associated dosage parameter. Here, we present a novel machine-learning approach towards learning counterfactual representations for estimating individual dose-response curves for any number of treatments with continuous dosage parameters with neural networks. Building on the established potential outcomes framework, we introduce performance metrics, model selection criteria, model architectures, and open benchmarks for estimating individual dose-response curves. Our experiments show that the methods developed in this work set a new state-of-the-art in estimating individual dose-response.
\end{abstract}
\section{Introduction}
Estimating dose-response curves from observational data is an important problem in many domains. In medicine, for example, we would be interested in using data of people that have been treated in the past to predict which treatments and associated dosages would lead to better outcomes for new patients \cite{imbens2000role}. This question is, at its core, a counterfactual one, i.e. we are interested in predicting what \textit{would have happened if} we were to give a patient a specific treatment at a specific dosage in a given situation. Answering such counterfactual questions is a challenging task that requires either further assumptions about the underlying data-generating process or prospective interventional experiments, such as randomised controlled trials (RCTs) \cite{stone1993assumptions,pearl2009causality,peters2017elements}. However, performing prospective experiments is expensive, time-consuming, and, in many cases, ethically not justifiable \cite{schafer1982ethics}. Two aspects make estimating counterfactual outcomes from observational data alone difficult \cite{yoon2018ganite,schwab2018pm}: Firstly, we only observe the factual outcome and never the counterfactual outcomes that would potentially have happened had we chosen a different treatment option. In medicine, for example, we only observe the outcome of giving a patient a specific treatment at a specific dosage, but we never observe what would have happened if the patient was instead given a potential alternative treatment or a different dosage of the same treatment. Secondly, treatments are typically not assigned at random in observational data. In the medical setting, physicians take a range of factors, such as the patient's expected response to the treatment, into account when choosing a treatment option. Due to this treatment assignment bias, the treated population may differ significantly from the general population. A supervised model na\"ively trained to minimise the factual error would overfit to the properties of the treated group, and therefore not generalise to the entire population.

To address these problems, we introduce a novel methodology for training neural networks for counterfactual inference that extends to any number of treatments with continuous dosage parameters. In order to control for the biased assignment of treatments in observational data, we combine our method with a variety of regularisation schemes originally developed for the discrete treatment setting, such as distribution matching \cite{johansson2016learning,shalit2016estimating}, propensity dropout (PD) \cite{alaa2017deep}, and matching on balancing scores \cite{rosenbaum1983central,ho2007matching,schwab2018pm}. In addition, we devise performance metrics, model selection criteria and open benchmarks for estimating individual dose-response curves. Our experiments demonstrate that the methods developed in this work set a new state-of-the-art in inferring individual dose-response curves. The source code for this work is available at \url{https://github.com/d909b/drnet}.

\paragraph{Contributions.} We present the following contributions:
\setlist{nolistsep}
\begin{itemize}[noitemsep,leftmargin=2.2ex]
\item We introduce a novel methodology for training neural networks for counterfactual inference that, in contrast to existing methods, is suitable for estimating counterfactual outcomes for any number of treatment options with associated exposure parameters.
\item We develop performance metrics, model selection criteria, model architectures, and open benchmarks for estimating individual dose-response curves.
\item We extend state-of-the-art methods for counterfactual inference for two non-parametric treatment options to the multiple parametric treatment options setting.
\item We perform extensive experiments that show that our method sets a new state-of-the-art in inferring individual dose-response curves from observational data across several challenging datasets.
\end{itemize}

\section{Related Work}
\label{sec:related_work}

\paragraph{Background.} Causal analysis of treatment effects with rigorous experiments is, in many domains, an essential tool for validating interventions. In medicine, prospective experiments, such as RCTs, are the de facto gold standard to evaluate whether a given treatment is efficacious in treating a specific indication across a population \cite{carpenter2014reputation,bothwell2016rct}. However, performing prospective experiments is expensive, time-consuming, and often not possible for ethical reasons \cite{schafer1982ethics}. Historically, there has therefore been considerable interest in developing methodologies for performing causal inference using readily available observational data \cite{granger1969investigating,angrist1996identification,rosenbaum1983central,robins2000marginal,pearl2009causality,hernan2016using,lake2017building}. The na\"ive approach of training supervised models to minimise the observed factual error is in general not a suitable choice for counterfactual inference tasks due to treatment assignment bias and the inability to observe counterfactual outcomes. To address the shortcomings of unsupervised and supervised learning in this setting, several adaptations to established machine-learning methods that aim to enable the estimation of counterfactual outcomes from observational data have recently been proposed \cite{johansson2016learning,shalit2016estimating,wager2017estimation,alaa2017bayesian,alaa2017deep,louizos2017causal,yoon2018ganite,schwab2018pm}. In this work, we build on several of these advances to develop a machine-learning approach for estimating individual dose-response with neural networks.

\paragraph{Estimating Individual Treatment Effects (ITE).}\footnote{The ITE is sometimes also referred to as the conditional average treatment effect (CATE).}  Matching methods \cite{ho2007matching} are among the most widely used approaches to causal inference from observational data. Matching methods estimate the counterfactual outcome of a sample $X$ to a treatment $t$ using the observed factual outcome of its nearest neighbours that have received $t$. Propensity score matching (PSM) \cite{rosenbaum1983central} combats the curse of dimensionality of matching directly on the covariates $X$ by instead matching on the scalar probability $p(t|X)$ of receiving a treatment $t$ given the covariates $X$. Another category of approaches uses adjusted regression models that receive both the covariates $X$ and the treatment $t$ as inputs. The simplest such model is Ordinary Least Squares (OLS), which may use either one model for all treatments, or a separate model for each treatment \cite{kallus2017recursive}. More complex models based on neural networks, like Treatment Agnostic Representation Networks (TARNETs), may be used to build non-linear regression models \cite{shalit2016estimating}. Estimators that combine a form of adjusted regression with a model for the exposure in a manner that makes them robust to misspecification of either are referred to as doubly robust \cite{funk2011doubly}. In addition to OLS and neural networks, tree-based estimators, such as Bayesian Additive Regression Trees (BART) \cite{chipman2010bart,chipman2016bayestree} and Causal Forests (CF) \cite{wager2017estimation}, and distribution modelling methods, such as Causal Multi-task Gaussian Processes (CMGP) \cite{alaa2017bayesian}, Causal Effect Variational Autoencoders (CEVAEs) \cite{louizos2017causal}, and Generative Adversarial Nets for inference of Individualised Treatment Effects (GANITE) \cite{yoon2018ganite}, have also been proposed for ITE estimation.\footnote{See \cite{knaus2018machine} and \cite{schwab2018pm} for empirical comparisons of large-numbers of machine-learning methods for ITE estimation for two and more available treatment options.} Other approaches, such as balancing neural networks (BNNs) \cite{johansson2016learning} and counterfactual regression networks (CFRNET) \cite{shalit2016estimating}, attempt to achieve balanced covariate distributions across treatment groups by explicitly minimising the empirical discrepancy distance between treatment groups using metrics such as the Wasserstein distance \cite{cuturi2013sinkhorn}. Most of the works mentioned above focus on the simplest setting with two available treatment options without associated dosage parameters. A notable exception is the generalised propensity score (GPS) \cite{imbens2000role} that extends the propensity score to treatments with continuous dosages.

In contrast to existing methods, we present the first machine-learning approach to learn to estimate individual dose-response curves for multiple available treatments with a continuous dosage parameter from observational data with neural networks. We additionally extend several known regularisation schemes for counterfactual inference to address the treatment assignment bias in observational data. To facilitate future research in this important area, we introduce performance metrics, model selection criteria, and open benchmarks. We believe this work could be particularly important for applications in precision medicine, where the current state-of-the-art of estimating the average dose response across the entire population does not take into account individual differences, even though large differences in dose-response between individuals are well-documented for many diseases \cite{zaske1982wide,oldenhof1988clinical,campbell2007aspirin}.

\section{Methodology}
\paragraph{Problem Statement.} We consider a setting in which we are given $N$ observed samples $X$ with $p$ pre-treatment covariates $x_i$ and $i \in [0 \isep p - 1]$. For each sample, the potential outcomes $y_{n,t}(s_t)$ are the response of the $n$th sample to a treatment $t$ out of the set of $k$ available treatment options $T = \{0, ..., k-1\}$ applied at a dosage $s_t \in \{s_t \in \mathbb{R}, a_t > 0 \text{ }|\text{ }a_t \leq s \leq b_t\}$, where $a_t$ and $b_t$ are the minimum and maximum dosage for treatment $t$, respectively. The set of treatments $T$ can have two or more available treatment options. As training data, we receive factual samples $X$ and their observed outcomes $y_{n,f}(s_f)$ after applying a specific observed treatment $f$ at dosage $s_f$. Using the training data with factual outcomes, we wish to train a predictive model to produce accurate estimates $\hat{y}_t(n, s)$ of the potential outcomes across the entire range of $s$ for all available treatment options $t$. We refer to the range of potential outcomes $y_{n,t}(s)$ across $s$ as the \textit{individual dose-response curve} of the $n$th sample. This setting is a direct extension of the Rubin-Neyman potential outcomes framework \cite{rubin2005causal}.

\paragraph{Assumptions.} Following \cite{imbens2000role,lechner2001identification}, we assume unconfoundedness, which consists of three key parts: (1) Conditional Independence Assumption: The assignment to treatment $t$ is independent of the outcome $y_t$ given the pre-treatment covariates $X$, (2) Common Support Assumption: For all values of $X$, it must be possible to observe all treatment options with a probability greater than 0, and (3) Stable Unit Treatment Value Assumption: The observed outcome of any one unit must be unaffected by the assignments of treatments to other units. In addition, we assume smoothness, i.e. that units with similar covariates $x_i$ have similar outcomes $y$, both for model training and selection.

\paragraph{Metrics.} To enable a meaningful comparison of models in the presented setting, we use metrics that cover several desirable aspects of models trained for estimating individual dose-response curves. Our proposed metrics respectively aim to measure a predictive model's ability (1) to recover the dose-response curve across the entire range of dosage values, (2) to determine the optimal dosage point for each treatment, and (3) to deduce the optimal treatment policy overall, including selection of the right treatment and dosage point, for each individual case. To measure to what degree a model covers the entire range of individual dose-response curves, we use the mean integrated square error\footnote{A normalised version of this metric has been used in \citet{silva2016observational}.} (MISE) between the true dose-response $y$ and the predicted dose-response $\hat{y}$ as estimated by the model over $N$ samples, all treatments $T$, and the entire range $[a_t, b_t]$ of dosages $s$.
\vskip -3.0ex
\begin{small}
\begin{align}
\text{MISE} = \frac{1}{N} \frac{1}{|T|} \sum_{t\in T}\sum_{n=1}^{N} \int_{s=a_t}^{b_t} \Big( y_{n,t}(s) - \hat{y}_{n,t}(s) \Big)^2 ds
\end{align}
\end{small}
\vskip -2ex{\setlength{\parindent}{0cm}\noindent
To further measure a model's ability to determine the optimal dosage point for each individual case, we calculate the mean dosage policy error (DPE). The mean dosage policy error is the mean squared error in outcome $y$ associated with using the estimated optimal dosage point $\hat{s}^{\ast}_{t}$ according to the predictive model to determine the \textit{true} optimal dosage point ${s}^{\ast}_{t}$ over $N$ samples and all treatments $T$.
}
\vskip -2ex
\begin{small}
\begin{align}
\text{DPE} &= \frac{1}{N} \frac{1}{|T|} \sum_{t\in T}\sum_{n=1}^{N} \Big( y_{n,t}({s}^{\ast}_{t}) - y_{n,t}(\hat{s}^{\ast}_{t})\Big)^2
\end{align}
\end{small}
\vskip -2ex{\setlength{\parindent}{0cm}\noindent
where ${s}^{\ast}_{t}$ and $\hat{s}^{\ast}_{t}$ are the optimal dosage point according to the true dose-response curve and the estimated dose-response curve, respectively.
}
\vskip -2ex
\begin{minipage}{.5\linewidth}
\begin{small}
\begin{align}
{s}^{\ast}_{t} &= \argmax_{s\in[a_t,b_t]} y_{n,t}(s)
\end{align}
\end{small}
\end{minipage}%
\begin{minipage}{.5\linewidth}
\begin{small}
\begin{align}
\hat{s}^{\ast}_{t} &= \argmax_{s\in[a_t,b_t]} \hat{y}_{n,t}(s)
\end{align}
\end{small}
\end{minipage}
\vskip -2.0ex
{\setlength{\parindent}{0cm}\noindent
Finally, the policy error (PE) measures a model's ability to determine the optimal treatment policy for individual cases, i.e. how much worse the outcome would be when using the estimated best optimal treatment option as opposed to the \textit{true} optimal treatment option and dosage.
}
\vskip -3.25ex
\begin{small}
\begin{align}
\text{PE} = \frac{1}{N}\sum_{n=1}^{N} \Big( y_{n,{t}^{\ast}}({s}^{\ast}_{{t}^{\ast}}) - y_{n,\hat{t}^{\ast}}(\hat{s}^{\ast}_{\hat{t}^{\ast}})\Big)^2 
\end{align}
\end{small}
\vskip -3.5ex{\setlength{\parindent}{0cm}\noindent
where \\
}
\vskip -5.5ex
\begin{minipage}{.5\linewidth}
\begin{small}
\begin{align}
{t}^{\ast} &= \argmax_{t \in T} y_{n,t}({s}^{\ast}_{t})
\end{align}
\end{small}
\end{minipage}%
\begin{minipage}{.5\linewidth}
\begin{small}
\begin{align}
\hat{t}^{\ast} &= \argmax_{t \in T} \hat{y}_{n,t}(\hat{s}^{\ast}_{t})
\end{align}
\end{small}
\end{minipage}
\vskip -2.5ex
{\setlength{\parindent}{0cm}\noindent
are the optimal treatment option according to the ground truth $y$ and the predictive model, respectively. Considering the DPE and PE alongside the MISE is important to comprehensively evaluate models for counterfactual inference. For example, a model that accurately recovers dose response curves outside the regions containing the optimal response would achieve a respectable MISE but would not be a good model for determining the treatment and dosage choices that lead to the best outcome for a given unit. By considering multiple metrics, we can ensure that predictive models are a capable both in recovering the entire dose-response as well as in selecting the best treatment and dosage choices. We note that, in general, we can not calculate the MISE, DPE or PE without knowledge of the outcome-generating process, since the true dose-response function $y_{n,t}(s)$ is unknown.
}

\begin{figure*}[t!]
\vskip -2.5ex
\centering
\includegraphics[width=0.95\linewidth]{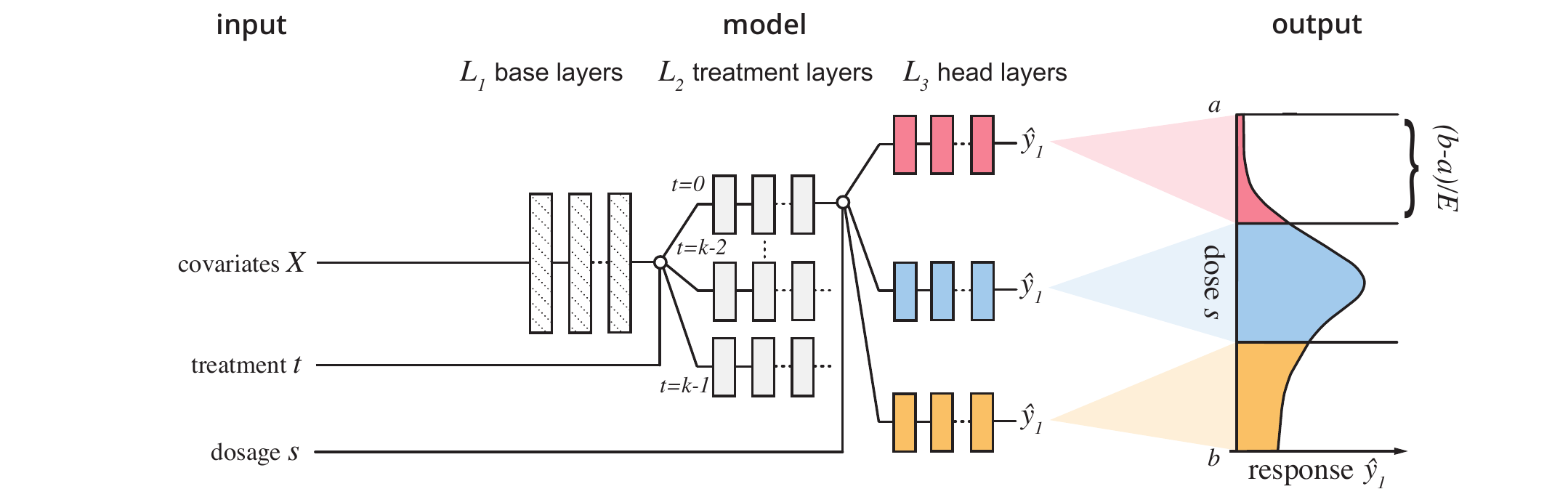}
\vskip -1.5ex
\caption{The dose response network (DRNet) architecture with shared base layers, $k$ intermediary treatment layers, and $k*E$ heads for the multiple treatment setting with an associated dosage parameter $s$. The shared base layers are trained on all samples, and the treatment layers are only trained on samples from their respective treatment category. Each treatment layer is further subdivided into $E$ head layers (only one set of $E=3$ head layers for treatment $t=0$ is shown above). Each head layer is assigned a dosage stratum that subdivides the range of potential dosages $[a_t, b_t]$ into $E$ partitions of equal width $(b-a)/E$. The head layers each predict outcomes $\hat{y}_t(s)$ for a range of values of the dosage parameter $s$, and are only trained on samples that fall within their respective dosage stratum. The hierarchical structure of DRNets enables them to share common hidden representations across all samples (base layers), treatment options (treatment layers), and dosage strata (head layers) while maintaining the influence of both $t$ and $s$ on the hidden layers.}
\label{fig:tarnet}
\vskip -3.5ex
\end{figure*}

\paragraph{Model Architecture.} Model structure plays an important role in learning representations for counterfactual inference with neural networks \cite{shalit2016estimating,schwab2018pm,alaa2018limits}. A particularly challenging aspect of training neural networks for counterfactual inference is that the influence of the treatment indicator variable $t$ may be lost in high-dimensional hidden representations \cite{shalit2016estimating}. To address this problem for the setting of two available treatments without dosage parameters, \citet{shalit2016estimating} proposed the TARNET architecture that uses a shared base network and separate head networks for both treatment options. In TARNETs, the head networks are only trained on samples that received the respective treatment. \citet{schwab2018pm} extended the TARNET architecture to the multiple treatment setting by using $k$ separate head networks, one for each treatment option. In the setting with multiple treatment options with associated dosage parameters, this problem is further compounded because we must maintain not only the influence of $t$ on the hidden representations throughout the network, but also the influence of the continuous dosage parameter $s$. To ensure the influence of both $t$ and $s$ on hidden representations, we propose a hierarchical architecture for multiple treatments called dose response network (DRNet, Figure \ref{fig:tarnet}). DRNets ensure that the dosage parameter $s$ maintains its influence by assigning a head to each of $E\in\mathbb{N}$ equally-sized dosage strata that subdivide the range of potential dosage parameters $[a_t, b_t]$. The hyperparameter $E$ defines the trade-off between computational performance and the resolution $\frac{(b-a)}{E}$ at which the range of dosage values is partitioned. To further attenuate the influence of the dosage parameter $s$ within the head layers, we additionally repeatedly append $s$ to each hidden layer in the head layers. We motivate the proposed hierarchical structure with the effectiveness of the regress and compare approach to counterfactual inference \cite{kallus2017recursive}, where one builds a separate estimator for each available treatment option. Separate models for each treatment option  suffer from data-sparsity, since only units that received each respective treatment can be used to train a per-treatment model and there may not be a large number of samples available for each treatment. DRNets alleviate the issue of data-sparsity by enabling information to be shared both across the entire range of dosages through the treatment layers and across treatments through the base layers. 

\paragraph{Model Selection.} Given multiple models, it is not trivial to decide which model would perform better at counterfactual tasks, since we in general do not have access to the true dose-response to calculate error metrics like the ones given above. We therefore use a nearest neighbour approximation of the MISE to perform model selection using held-out factual data that has not been used for training. We calculate the nearest neighbour approximation NN-MISE of the MISE using:
\begin{small}
\begin{align}
\text{NN-MISE} = \frac{1}{N} \frac{1}{T} \sum_{t=1}^{T}\sum_{n=1}^{N} \int_{s=a_t}^{b_t} \Big( y_{\text{NN}(n),t}(s) - \hat{y}_{n,t}(s) \Big)^2 ds
\end{align}
\end{small}
where we substitute the true dose-response $y_{n,t}$ of the $n$th sample with the outcome $y_{\text{NN}(n),t}$ of an observed factual nearest neighbour of the $n$th sample at a dosage point $s$ from the training set. Using the nearest neighbour approximation of the MISE, we are able to perform model selection without access to the true counterfactual outcomes $y$. Among others, nearest neighbour methods have also been proposed for model selection in the setting with two available treatments without dosages \cite{schuler2018comparison}.

\paragraph{Regularisation Schemes.} DRNets can be combined with regularisation schemes developed to further address treatment assignment bias. To determine the utility of various regularisation schemes, we evaluated DRNets using distribution matching \cite{shalit2016estimating}, propensity dropout \cite{alaa2017deep}, matching on the entire dataset \cite{ho2007matching}, and on the batch level \cite{schwab2018pm}. We na\"ively extended these regularisation schemes since neither of these methods were originally developed for the dose-response setting (Appendix A).

\section{Experiments}
Our experiments aimed to answer the following questions:
\setlist{nolistsep}
\begin{itemize}[noitemsep,leftmargin=1ex]
\item[1] How does the performance of our proposed approach compare to state-of-the-art methods for estimating individual dose-response?
\item[2] How do varying choices of $E$ influence counterfactual inference performance?
\item[3] How does increasing treatment assignment bias affect the performance of dose-response estimators?
\end{itemize}

\paragraph{Datasets.} Using real-world data, we performed experiments on three semi-synthetic datasets with two and more treatment options to gain a better understanding of the empirical properties of our proposed approach. To cover a broad range of settings, we chose datasets with different outcome and treatment assignment functions, and varying numbers of samples, features and treatments (Table \ref{tb:datasets}). All three datasets were randomly split into training (63\%), validation (27\%) and test sets (10\%).

\paragraph{News.} The News benchmark consisted of 5000 randomly sampled news articles from the NY Times corpus\footnote{https://archive.ics.uci.edu/ml/datasets/bag+of+words} and was originally introduced as a benchmark for counterfactual inference in the setting with two treatment options without an associated dosage parameter \cite{johansson2016learning}. We extended the original dataset specification \cite{johansson2016learning,schwab2018pm} to enable the simulation of any number of treatments with associated dosage parameters. The samples $X$ were news articles that consist of word counts $x_i \in \mathbb{N}$, outcomes $y_{s,t} \in \mathbb{R}$ that represent the reader's opinion of the news item, and a normalised dosage parameter $s_t \in (0,1]$ that represents the viewer's reading time. There was a variable number of available treatment options $t$ that corresponded to various devices that could be used to view the News items, e.g. smartphone, tablet, desktop, television or others \cite{johansson2016learning}. We trained a topic model on the entire NY Times corpus to model that consumers prefer to read certain media items on specific viewing devices. We defined $z(X)$ as the topic distribution of news item $X$, and  randomly picked $k$ topic space centroids $z_t$ and $2k$ topic space centroids $z_{s_t,i}$ with $i\in{0,1}$ as prototypical news items. We assigned a random Gaussian outcome distribution with mean $\mu \sim \mathcal{N}(0.45, 0.15)$ and standard deviation $\sigma \sim \mathcal{N}(0.1,0.05)$ to each centroid. For each sample, we drew ideal potential outcomes from that Gaussian outcome distribution $\tilde{y}_t \sim \mathcal{N}(\mu_t, \sigma_t) + \epsilon$ with $\epsilon \sim \mathcal{N}(0, 0.15)$. The dose response $\tilde{y}_s$ was drawn from a distance-weighted mixture of two Gaussians $\tilde{y}_s \sim d_0 \mathcal{N}(\mu_{s_t,0}, \sigma_{s_t,0}) + d_1 \mathcal{N}(\mu_{s_t,1}, \sigma_{s_t,1})$ using topic space distances $d=\text{softmax}(\text{D}(z(X), z_{s_t,i}))$ and the Euclidean distance as distance metric $\text{D}$. We assigned the observed treatment $t$ using $t|x \sim \text{Bern}(\text{softmax}(\kappa \tilde{y}_t \tilde{y}_s))$ with a treatment assignment bias coefficient $\kappa$ and an exponentially distributed observed dosage $s_t$ using $s_t \sim \text{Exp}(\beta)$ with $\beta = 0.25$. The true potential outcomes $y_{s,t} = C\tilde{y}_t \tilde{y}_s$ were the product of $\tilde{y}_t$ and $\tilde{y}_s$ scaled by a coefficient $C=50$. We used four different variants of this dataset with $k=2$, $4$, $8$, and $16$ viewing devices, and $\kappa = 10$, $10$, $10$, and $7$, respectively. Higher values of $\kappa$ indicate a higher expected treatment assignment bias depending on $\tilde{y}_t \tilde{y}_s$, with $\kappa = 0$ indicating no assignment bias. 

\paragraph{Mechanical Ventilation in the Intensive Care Unit (MVICU).} The MVICU benchmark models patients' responses to different configuratations of mechanical ventilation in the intensive care unit. The data was sourced from the publicly available MIMIC III database \cite{saeed2011multiparameter}. The samples $X$ consisted of the last observed measurements $x_i$ of various biosignals, including respiratory, cardiac and ventilation signals. The outcomes were arterial blood gas readings of the ratio of arterial oxygen partial pressure to fractional inspired oxygen $PaO_2 / FiO_2$ which, at values lower than 300, are used as one of the clinical criteria for the diagnosis Acute Respiratory Distress Syndrome (ARDS) \cite{ferguson2012berlin}. We modelled a mechanical ventilator with $k=3$ adjustable treatment parameters: (1) the fraction of inspired oxygen, (2) the positive end-expiratory pressure in the lungs, and (3) tidal volume. To model the outcomes, we use the same procedure as for the News benchmark with a Gaussian outcome function and a mixture of Gaussian dose-response function, with the exception that we did not make use of topic models and instead performed the similarity comparisons D in covariate space. We used a treatment assignment bias $\kappa=10$ and a scaling coefficient $C=150$. Treatment dosages were drawn according to $s_t \sim \mathcal{N}(\mu_{\text{dose},t}, 0.1)$, where the distribution means were defined as $\mu_{\text{dose}} = (0.6,0.65, 0.4)$ for each treatment.

\paragraph{The Cancer Genomic Atlas (TCGA).} The TCGA project collected gene expression data from various types of cancers in 9659 individuals \cite{weinstein2013cancer}. There were $k=3$ available clinical treatment options: (1) medication, (2) chemotherapy, and (3) surgery. We used a synthetic outcome function that simulated the risk of cancer recurrence after receiving either of the treatment options based on the real-world gene expression data. We standardised the gene expression data using the mean and standard deviations of gene expression at each gene locus for normal tissue in the training set. To model the outcomes,  we followed the same approach as in the MVICU benchmark with similarity comparisons done in covariate space using the cosine similarity as distance metric D, and parameterised with $\kappa=10$ and $C=50$. Treatment dosages in the TCGA benchmark were drawn according to $s_t \sim \mathcal{N}(0.65, 0.1)$.

\begin{minipage}{\textwidth}
\centering
\begin{minipage}[b]{.47\textwidth}
\centering
\begin{small}
\begin{tabular}{l@{\hskip 1.5ex}r@{\hskip 1.8ex}r@{\hskip 1.8ex}r@{\hskip 1.8ex}r}
\toprule
Dataset & \# Samples & \# Features & \# Treatments \\
\midrule
News & 5000 & 2870 & 2/4/8/16 \\
MVICU & 8040 & 49 & 3 \\
TCGA & 9659 & 20531 & 3 \\
\bottomrule
\end{tabular}
\end{small}
\captionof{table}{Comparison of the benchmark datasets used in our experiments. We evaluate on three semi-synthetic datasets with varying numbers of treatments and samples.}
\label{tb:datasets}
\end{minipage}
\hfill
\begin{minipage}[b]{.47\textwidth}
  \centering
  \includegraphics[width=0.98\linewidth]{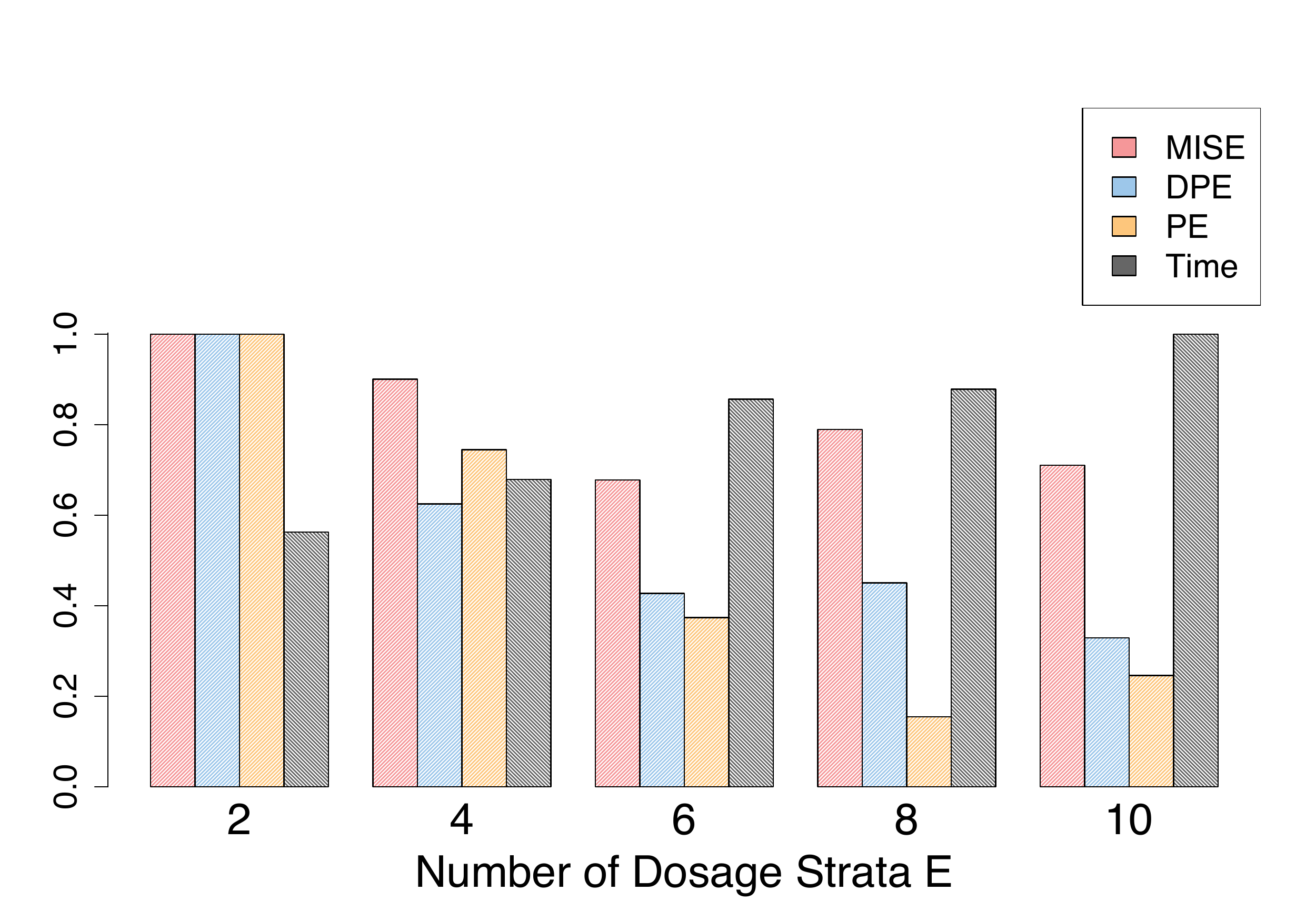}
  \captionof{figure}{Analysis of the effect of choosing various numbers of dosage strata $E$ (x-axis) on MISE (red), DPE (blue), PE (orange) and Time needed for training and evaluation (black) as calculated on the MVICU benchmark. Metrics were normalised to the range $[0, 1]$. All other hyperparameters besides $E$ were held equal.} 
  \label{fig:strata}
\end{minipage}
\end{minipage}

\paragraph{Models.} We evaluated DRNet, ablations, baselines, and all relevant state-of-the-art methods: k-nearest neighbours (kNN) \cite{ho2007matching}, BART \cite{chipman2010bart,chipman2016bayestree}, CF \cite{wager2017estimation}, GANITE \cite{yoon2018ganite}, TARNET \cite{shalit2016estimating}, and GPS \cite{imbens2000role} using the "causaldrf" package \cite{galagate2016causal}. We evaluated which regularisation strategy for learning counterfactual representations is most effective by training DRNets using a Wasserstein regulariser between treatment group distributions (+ Wasserstein) \cite{shalit2016estimating}, PD (+ PD) \cite{alaa2017deep}, batch matching (+ PM) \cite{schwab2018pm}, and matching the entire training set as a preprocessing step \cite{ho2011matchit} using the PM algorithm (+ PSM$_\text{PM}$) \cite{schwab2018pm}. To determine whether the DRNet architecture is more effective than its alternatives at learning representations for counterfactual inference in the presented setting, we also evaluated (1) a multi-layer perceptron (MLP) that received the treatment index $t$ and dosage $s$ as additional inputs, and (2) a TARNET for multiple treatments that received the dosage $s$ as an extra input (TARNET) \cite{johansson2016learning,schwab2018pm} with all other hyperparameters beside the architecture held equal. As a final ablation of DRNet, we tested whether appending the dosage parameter $s$ to each hidden layer in the head networks is effective by also training DRNets that only receive the dosage parameter once in the first hidden layer of the head network (- Repeat). We na\"ively extended CF, GANITE and BART by adding the dosage as an additional input covariate, because they were not designed for treatments with dosages.

\paragraph{Hyperparameters.} To ensure a fair comparison of the tested models, we took a systematic approach to hyperparameter search. Each model was given exactly the same number of hyperparameter optimisation runs with hyperparameters chosen at random from predefined hyperparameter ranges (Appendix B). We used 5 hyperparameter optimisation runs for each model on TCGA and 10 on all other benchmarks. Furthermore, we used the same random seed for each model, i.e. all models were evaluated on exactly the same sets of hyperparameter configurations. After computing the hyperparameter runs, we chose the best model based on the validation set NN-MISE. This setup ensures that each model received the same degree of hyperparameter optimisation. For all DRNets and ablations, we used $E=5$ dosage strata with the exception of those presented in Figure \ref{fig:strata}.
\begin{figure}[b!]
\vskip -3.5ex
\centering
\includegraphics[width=.38\linewidth]{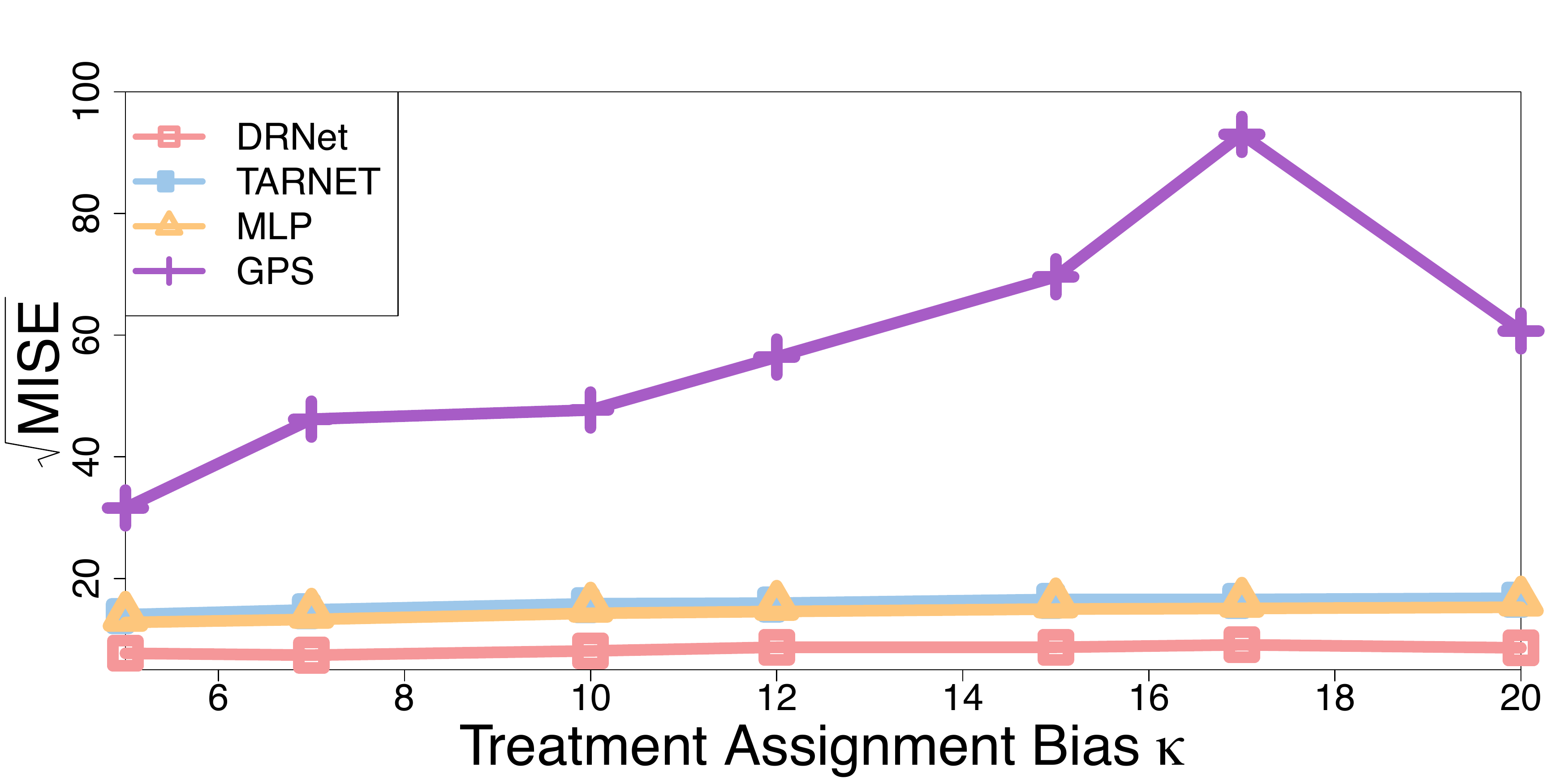}\hspace{6ex}
\includegraphics[width=.38\linewidth]{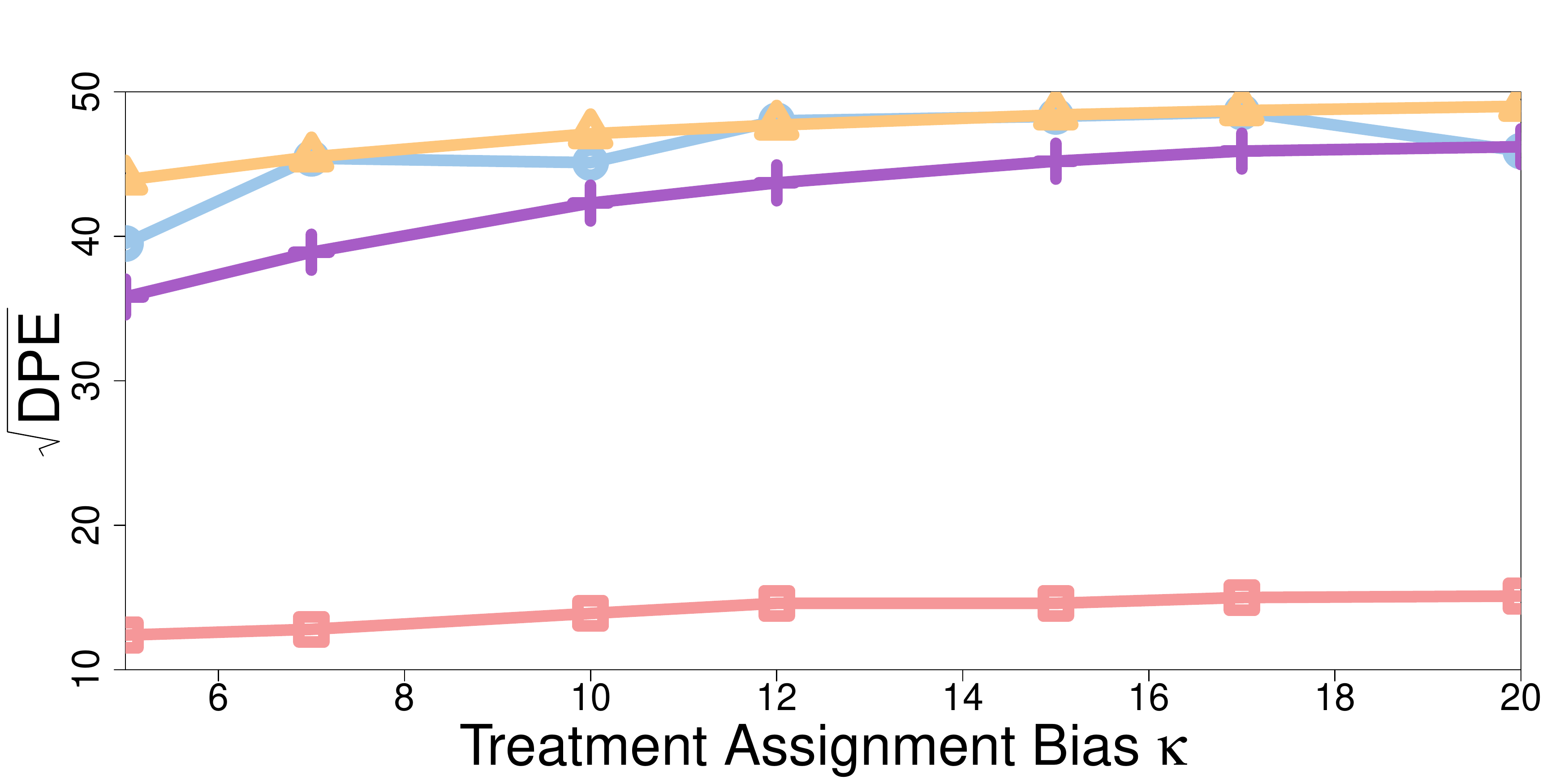}
\vskip -0.75ex
\caption{Comparison of DRNet (red), TARNET (blue), MLP (yellow) and GPS (purple) in terms of their $\sqrt{\text{MISE}}$ (top) and $\sqrt{\text{DPE}}$ (bottom) for varying levels of treatment assignment bias $\kappa$ (x-axis) on News-2. DRNet performs better than other methods across the entire evaluated range of treatment assignment bias values, and is more robust to increasing levels of $\kappa$.}
\label{fig:kappa}
\vskip -1.5ex
\end{figure}
\paragraph{Metrics.} For each dataset and model, we calculated the $\sqrt{\text{MISE}}$, $\sqrt{\text{DPE}}$, and $\sqrt{\text{PE}}$. We used Romberg integration with $64$ equally spaced samples from $y_{n,t}$ and $\hat{y}_{n,t}$ to compute the inner integral over the range of dosage parameters necessary for the MISE metric. To compute the optimal dosage points and treatment options in the DPE and PE, we used Sequential Least Squares Programming (SLSQP) to determine the respective maxima of $y_{n,t}(s)$ and $\hat{y}_{n,t}(s)$ numerically.

\begin{table*}[t!]
\vspace{-1.5ex}
\setlength{\tabcolsep}{1.38ex}
\caption{Comparison of methods for counterfactual inference with multiple parametric treatments on News-2/4/8/16, MVICU and TCGA. We report the mean value $\pm$ the standard deviation of $\sqrt{\text{MISE}}$ on the respective test sets over 5 repeat runs with new random seeds. n.r. = not reported for computational reasons (excessive runtime). \sig = significantly different from DRNet ($\alpha < 0.05$).}
\label{tb:results_1}
\centering
\begin{small}
\begin{tabular}{lrrrrrr}
\toprule
Method & News-2 & News-4 & News-8 & News-16 & MVICU & TCGA \\
\midrule
{DRNet}  &  {8.0} $\pm$ 0.1 & {11.6} $\pm$ 0.1 & {10.2} $\pm$ 0.1 & {10.3} $\pm$ 0.0 & {31.1} $\pm$ 0.4 & \textbf{9.6} $\pm$ 0.0  \\
- Repeat  &  \sig\hspace{1ex}{9.0} $\pm$ 0.1 & \sig{11.9} $\pm$ 0.2 & {10.3} $\pm$ 0.1 & {10.4} $\pm$ 0.1 & \textbf{31.0} $\pm$ 0.3 & {10.2} $\pm$ 0.2  \\
\midrule
+ Wasserstein  &  \sig\hspace{1ex}\textbf{7.7} $\pm$ 0.2 & \textbf{11.5} $\pm$ 0.0 & \sig\textbf{10.0} $\pm$ 0.0 & \sig\textbf{10.2} $\pm$ 0.0 & {32.9} $\pm$ 2.9 & {10.2} $\pm$ 0.9  \\
+ PD  &  \sig\hspace{1ex}{9.0} $\pm$ 0.2 & \sig{12.2} $\pm$ 0.1 & \sig{10.6} $\pm$ 0.2 & {10.3} $\pm$ 0.1 & \sig{36.9} $\pm$ 0.9 & \sig{11.9} $\pm$ 1.4  \\
+ PM  &  \sig\hspace{1ex}{8.4} $\pm$ 0.3 & \sig{12.2} $\pm$ 0.1 & \sig{11.4} $\pm$ 0.3 & \sig{12.3} $\pm$ 0.3 & {31.2} $\pm$ 0.4 & {9.7} $\pm$ 0.2  \\
+ PSM$_\text{PM}$  &  \sig\hspace{1ex}{8.6} $\pm$ 0.1 & \sig{12.2} $\pm$ 0.2 & \sig{11.5} $\pm$ 0.2 & \sig{12.2} $\pm$ 0.3 & \sig{32.6} $\pm$ 0.5 & \sig{11.4} $\pm$ 0.6  \\
\midrule
MLP  &  \sig{15.3} $\pm$ 0.1 & \sig{14.5} $\pm$ 0.0 & \sig{13.9} $\pm$ 0.1 & \sig{14.0} $\pm$ 0.0 & \sig{49.5} $\pm$ 5.1 & \sig{15.3} $\pm$ 0.2  \\
TARNET  &  \sig{15.5} $\pm$ 0.1 & \sig{15.4} $\pm$ 0.0 & \sig{14.7} $\pm$ 0.1 & \sig{14.7} $\pm$ 0.1 & \sig{58.0} $\pm$ 4.8 & \sig{14.7} $\pm$ 0.1  \\
GANITE  &  \sig{16.8} $\pm$ 0.1 & \sig{15.6} $\pm$ 0.1 & \sig{14.8} $\pm$ 0.1 & \sig{14.8} $\pm$ 0.0 & \sig{59.5} $\pm$ 0.8 & \sig{15.4} $\pm$ 0.2  \\
\midrule
kNN  &  \sig{16.2} $\pm$ 0.0 & \sig{14.7} $\pm$ 0.0 & \sig{15.0} $\pm$ 0.0 & \sig{14.5} $\pm$ 0.0 & \sig{54.9} $\pm$ 0.0 & n.r.  \\
GPS  &  \sig{47.6} $\pm$ 0.1 & \sig{24.7} $\pm$ 0.1 & \sig{22.9} $\pm$ 0.0 & \sig{15.5} $\pm$ 0.1 & \sig{78.3} $\pm$ 0.0 & \sig{26.3} $\pm$ 0.0  \\
\midrule
CF  &  \sig{26.0} $\pm$ 0.0 & \sig{20.5} $\pm$ 0.0 & \sig{19.6} $\pm$ 0.0 & \sig{14.9} $\pm$ 0.0 & \sig{57.5} $\pm$ 0.0 & \sig{15.2} $\pm$ 0.0  \\
BART  &  \sig{13.8} $\pm$ 0.2 & \sig{14.0} $\pm$ 0.1 & \sig{13.0} $\pm$ 0.1 & n.r. & \sig{47.1} $\pm$ 0.8 & n.r.  \\
\bottomrule
\end{tabular}
\end{small}
\vskip -2.75ex
\end{table*}

\section{Results and Discussion}

\paragraph{Counterfactual Inference.} In order to evaluate the relative performances of the various methods across a wide range of settings, we compared the MISE of the listed models for counterfactual inference on the News-2/4/8/16, MVICU and TCGA benchmarks (Table \ref{tb:results_1}; other metrics in Appendix D). Across the benchmarks, we found that DRNets outperformed all existing state-of-the-art methods in terms of MISE. We also found that DRNets that used additional regularisation strategies outperformed vanilla DRNets on News-2, News-4, News-8 and News-16. However, on MVICU and TCGA, DRNets that used additional regularisation performed similarly as standard DRNets. Where regularisation was effective, Wasserstein regularisation between treatment groups (+ Wasserstein) and batch matching (+ PM) were generally slightly more effective than PSM$_\text{PM}$ and PD. In addition, not repeating the dosage parameter for each layer in the per-dosage range heads of a DRNet (- Repeat) performed worse than appending the dosage parameter on News-2, News-4 and News-8. Lastly, the results showed that DRNet improved upon both TARNET and the MLP baseline by a large margin across all datasets - demonstrating that the hierarchical dosage subdivision introduced by DRNets is effective, and that an optimised model structure is paramount for learning representations for counterfactual inference. 

\paragraph{Number of Dosage Strata $E$.} To determine the impact of the choice of the number of dosage strata $E$ on DRNet performance, we analysed the estimation performance and computation time of DRNets trained with various numbers of dosage strata $E$ on the MVICU benchmark (Figure \ref{fig:strata}). With all other hyperparameters held equal, we found that a higher number of dosage strata in general improves estimation performance, because the resolution at which the dosage range is partitioned is increased. However, there is a trade-off between resolution and computational performance, as higher values of $E$ consistently increased the computation time necessary for training and prediction. 

\paragraph{Treatment Assignment Bias.} To assess the robustness of DRNets and existing methods to increasing levels of treatment assignment bias in observational data, we compared the performance of DRNet to TARNET, MLP and GPS on the test set of News-2 with varying choices of treatment assignment bias $\kappa \in [5, 20]$ (Figure \ref{fig:kappa}). We found that DRNet  outperformed existing methods across the entire range of evaluated treatment assignment biases. 

\paragraph{Limitations.} A general limitation of methods that attempt to estimate causal effects from observational data is that they are based on untestable assumptions \cite{stone1993assumptions}. In this work, we assume unconfoundedness \cite{imbens2000role,lechner2001identification}, which implies that one must have reasonable certainty that the available covariate set $X$ contains the most relevant variables for the problem setting being modelled. Making this judgement can be difficult in practice, particularly when one does not have much prior knowledge about the underlying causal process. Even without such certainty, this approach may nonetheless be a justifiable starting point to generate hypotheses when experimental data is not available \cite{imbens2004nonparametric}.

\section{Conclusion}
We presented a deep-learning approach to learning to estimate individual dose-response to multiple treatments with continuous dosage parameters based on observational data. We extended several existing regularisation strategies to the setting with any number of treatment options with associated dosage parameters, and combined them with our approach in order to address treatment assignment bias inherent in observational data. In addition, we introduced performance metrics, model selection criteria, model architectures, and new open benchmarks for this setting. Our experiments demonstrated that model structure is paramount in learning neural representations for counterfactual inference of dose-response curves from observational data, and that there is a trade-off between model resolution and computational performance in DRNets. DRNets significantly outperform existing state-of-the-art methods in inferring individual dose-response curves across several benchmarks. 

\subsubsection*{Acknowledgements}
This work was partially funded by the Swiss National Science Foundation (SNSF) project No. 167302 within the National Research Program (NRP) $75$ ``Big Data''. We gratefully acknowledge the support of NVIDIA Corporation with the donation of the Titan Xp GPUs used for this research. The results shown here are in whole or part based upon data generated by the TCGA Research Network: \url{http://cancergenome.nih.gov/}. 

\small
\bibliography{references}
\bibliographystyle{unsrtnat}

\includepdf[pages=1-]{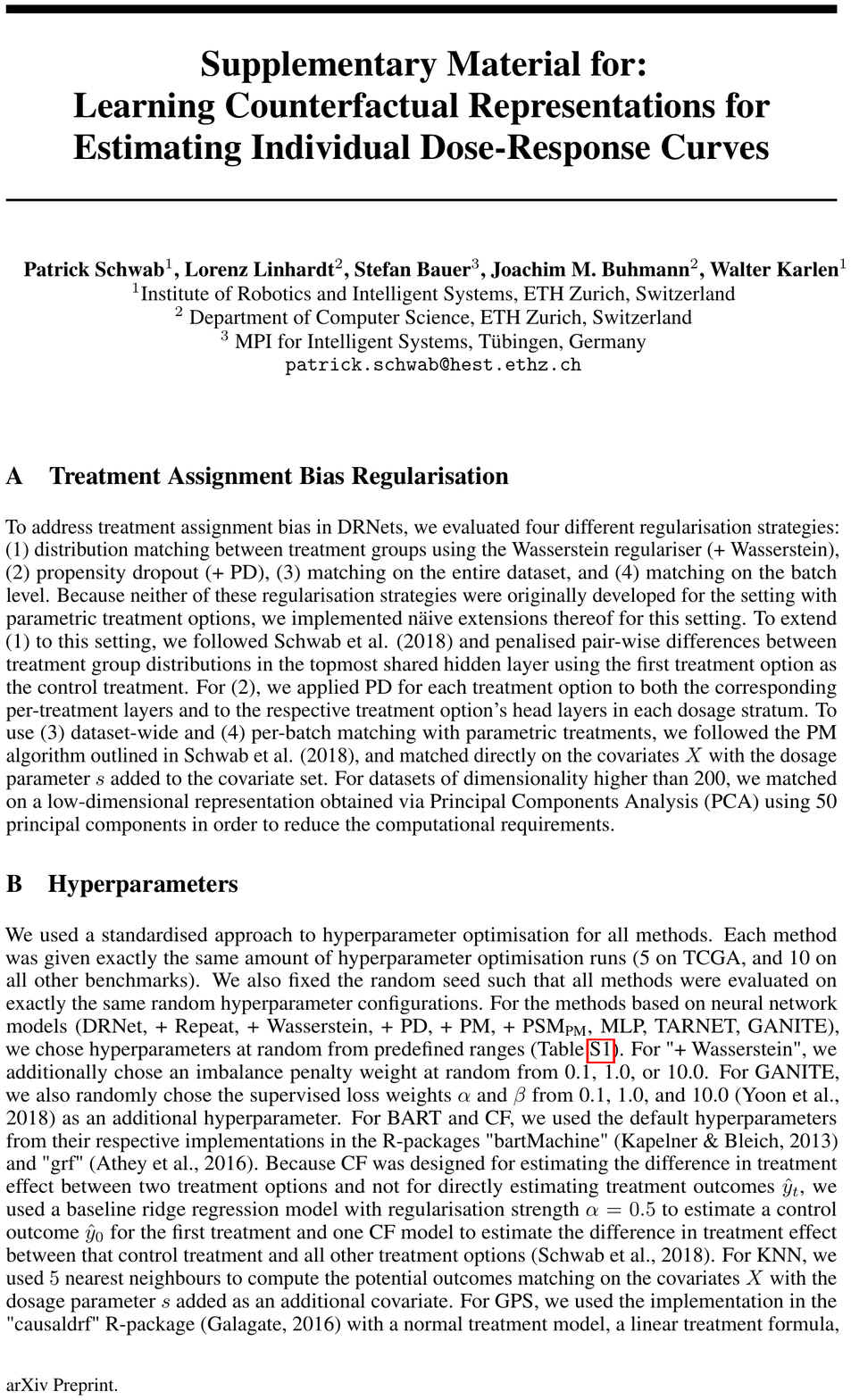}

\end{document}